\documentclass[final,1p,times]{elsarticle}

\usepackage[numbers]{natbib}
\usepackage{float}
\usepackage{graphicx}
\usepackage{verbatim}
\usepackage{appendix}
\usepackage{array}
\usepackage{multirow}
\usepackage{booktabs} 
\usepackage{setspace}
\restylefloat{table}

\usepackage{orcidlink}      
\usepackage{hyperref}       
\usepackage{fnpos}          

\usepackage{hyperref}
\hypersetup{
 colorlinks=true,
 citecolor=Blue,
 linkcolor=Blue,
 urlcolor=Blue}
\usepackage{url} 
\usepackage{amssymb}

\usepackage{amsmath}

\makeatletter
\def\ps@pprintTitle{%
  \let\@oddhead\@empty%
  \let\@evenhead\@empty%
  \let\@oddfoot\@empty%
  \let\@evenfoot\@empty}
\makeatother

\begin{document}

\begin{frontmatter}






\title{\textit{EvalYaks}: Instruction Tuning Datasets and LoRA Fine-tuned Models for Automated Scoring of CEFR B2 Speaking Assessment Transcripts}



\author[inst1,inst3]{Nicy Scaria\corref{cor1}\fnref{equal}\orcidlink{0009-0004-8699-0312}}
\ead{nicyscaria@iisc.ac.in}

\author[inst2,inst3]{Silvester John Joseph Kennedy\fnref{equal}}
\author[inst2,inst3]{Thomas Latinovich}
\author[inst1]{Deepak Subramani\orcidlink{0000-0002-5972-8878}}

\address[inst1]{Computational and Data Science, IISc, Bangalore, India}
\address[inst2]{Talking Yak, Inc., Cedarburg, Wisconsin, USA}
\address[inst3]{Talking Yak English Learning Private Limited, Bangalore, India}

\cortext[cor1]{Corresponding author: nicyscaria@iisc.ac.in}
\fntext[equal]{These authors contributed equally to this work.}











\begin{abstract}
Relying on human experts to evaluate CEFR speaking assessments in an e-learning environment creates scalability challenges, as it limits how quickly and widely assessments can be conducted. We aim to automate the evaluation of CEFR B2 English speaking assessments in e-learning environments from conversation transcripts. First, we evaluate the capability of leading open source and commercial Large Language Models (LLMs) to score a candidate's performance across various criteria in the CEFR B2 speaking exam in both global and India-specific contexts. Next, we create a new expert-validated, CEFR-aligned synthetic conversational dataset with transcripts that are rated at different assessment scores. In addition, new instruction-tuned datasets are developed from the English Vocabulary Profile (up to CEFR B2 level) and the CEFR-SP WikiAuto datasets. Finally, using these new datasets, we perform parameter efficient instruction tuning of Mistral Instruct 7B v0.2 to develop a family of models called \textit{EvalYaks}. Four models in this family are for assessing the four sections of the CEFR B2 speaking exam, one for identifying the CEFR level of vocabulary and generating level-specific vocabulary, and another for detecting the CEFR level of text and generating level-specific text. \textit{EvalYaks} achieved an average acceptable accuracy of 96\%, a degree of variation of 0.35 levels, and performed 3 times better than the next best model. This demonstrates that a 7B parameter LLM instruction tuned with high-quality CEFR-aligned assessment data can effectively evaluate and score CEFR B2 English speaking assessments, offering a promising solution for scalable, automated language proficiency evaluation.  
\end{abstract}




\end{frontmatter}

\section{Introduction}
\label{intro}

The study of English is classified into Academic and Functional (General) English, each fulfilling different use cases. Academic English, prevalent in professional and educational spheres such as universities, prioritizes a formal tone, organized writing, and precise vocabulary for tasks such as essays, reports, and scholarly communication. In contrast, Functional English targets everyday communication skills in speaking, listening, reading, and writing, aiming for practical application in social and personal interactions with a more informal, conversational approach \citep{north2010core}.

The Common European Framework of Reference for Languages (CEFR) evaluates English proficiency on a six-level scale from A1 (Beginner) to C2 (Advanced). A CEFR B2 qualification indicates that the learner has the ability to independently use English to live, work, or study \citep{council2001common}. The CEFR uses `can do' descriptors to tailor teaching and assessment, aligning curriculum and educational objectives. These descriptors help educators set communicative goals and adapt courses to specific learning needs through consultations with experts and stakeholders \citep{cefr_companion}. Learners often have a `spiky profile', excelling in some language skills but struggling in others, reflecting their Target Language Use (TLU). For example, the IELTS Life Skills Test \citep{ielts} assesses only speaking and listening skills for UK visa applicants. The CEFR framework requires adaptation to fit specific contexts and is not a universal solution. Its `can do' statements define TLUs such as Personal, Public, Occupational, and Educational, facilitating customized teaching and assessment \citep{cambridgeenglish2018cefr}. E-learning can create a comprehensive curriculum customized to individual preferences and abilities, effectively using the CEFR framework to meet specific needs of the learner and contextual demands.

Various organizations, including Cambridge English and the British Council, offer English language programs and assessments. Cambridge English conducts in-person exams in three categories: Schools, General and Higher Education, and Business. The General and Higher Education exams cater to career and academic requirements at five levels: A2 Key, B1 Preliminary, B2 First, C1 Advanced, and C2 Proficiency. The B2 First exam is crucial for showcasing communication skills in English-speaking environments. On the other hand, the British Council's EnglishScore \citep{englishscore2023validity} provides a more straightforward, mobile-based online assessment for general English users.

Our objective is to develop a range of models to automate the evaluation and scoring process with the ability to handle the complexities of advanced language examinations such as the B2 First. Using both international and India-specific data, we strive to improve the accuracy and relevance of assessments, particularly for Indian students, in both global and local contexts.

E-learning is uniquely positioned to address the preferences and capabilities of learners, particularly as the complexity of subjects increases \citep{morris2008economies}. It is especially beneficial for subjects that require in-depth knowledge and detailed responses. In this context, depending on human experts to evaluate every assessment can be extremely costly and does not scale efficiently \citep{shah2014some}. The adoption of technologies that can emulate the expertise of human assessors and automate the evaluation process offers a feasible solution \citep{mekterovic2023scaling}. Implementing technologies that replicate human expertise and automate evaluations could provide a solution, facilitating effective and unbiased teaching and assessment of complex topics, thereby making quality education more scalable and accessible.

Artificial intelligence (AI), including Generative AI (GenAI) technology, has made significant advancements in various domains, including education, healthcare, and scientific research. GenAI models \citep{genai} can produce diverse content such as text, images, and videos using AI techniques. The integration of AI in education is increasing, albeit at a slower pace compared to other industries, yet its potential influence on education is substantial \citep{chatgpt_education}. These technologies enable personalized educational materials, assessments, and tutoring \citep{ji2023systematic}. Research \citep{AIassistance} has indicated that incorporating AI into educational settings has the potential to enhance students' autonomy in managing their own learning processes. Conversational AI, which conducts human-like conversations via text or audio based on LLMs, is widely accepted among students for task-oriented dialogues \citep{ji2023systematic}. Tools like Google Assistant have improved EFL students' communication skills and attitudes towards intelligent assistants in learning \citep{tai2023impact}. The integration of LLMs with educational technology is being applied to a wide range of tasks, including automated grading of short answers \citep{schneider2023towards, kortemeyer2023performance}, essay scoring \citep{stahl-etal-2024-exploring}, generating questions automatically \citep{mulla2023automatic, scaria2024automated, scaria2024good}, creating multiple-choice questions \citep{feng2024exploring, lee2024math}, and developing chatbots \citep{dan2023educhat}. Chatbots providing feedback have effectively boosted vocabulary learning among Korean EFL primary students \citep{jeon2023chatbot}. GPT-4 has shown better performance over human responses in open-ended assessments at various cognitive levels, highlighting its potential to support advanced educational evaluations \citep{rodrigues2024assessing}.

For a long time, researchers have been developing automated language assessments to efficiently and accurately evaluate the abilities of English learners. SpeechRater$^\text{SM}$ Version 1.0 (v1.0) \citep{xi2008automated} is an automated system created to score the spontaneous speech of English learners, and it is operationally used in the Test of English as a Foreign Language$^\text{TM}$ (TOEFL\textregistered) Practice Online assessment. Students find automated scoring of speaking performance and feedback to be beneficial \citep{gu2021using}. Recent advancements include the implementation of transformer-based models like BERT \citep{rama2021pre} for evaluating CEFR levels of sentences, and GPT variants \citep{gpt4} for assessing essays written by L2 English learners \citep{yancey2023rating}. These tools aim to provide unbiased, globally applicable evaluations and focus on aligning content with recognized proficiency levels. 

\begin{figure}[h!]
	\centering
		\includegraphics[width=\textwidth]{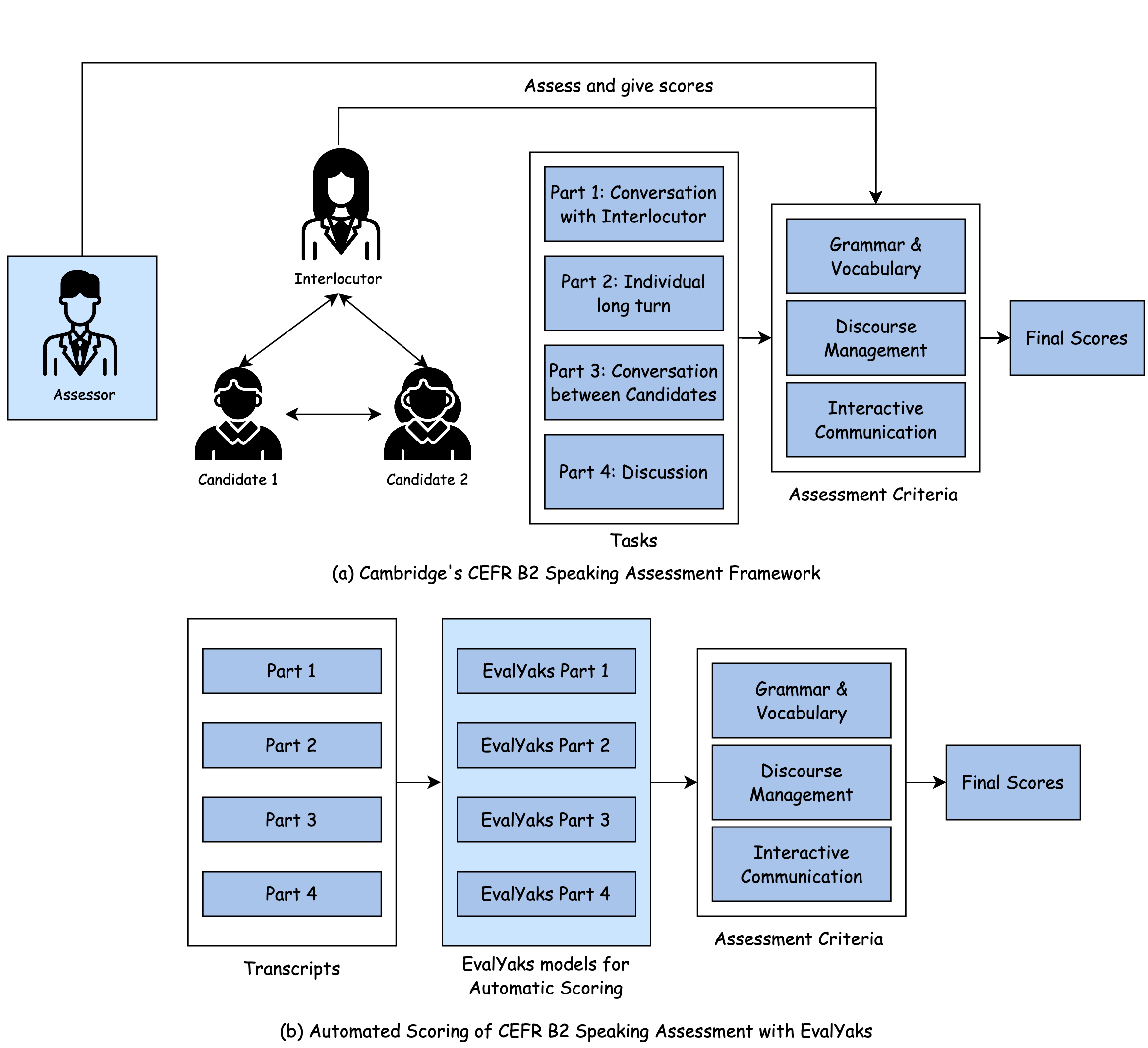}
	  \caption{Cambridge's CEFR B2 Speaking Assessment framework and it's automated scoring using \textit{EvalYaks.}}\label{fig:blockdiagram}
\end{figure}

The primary aim of this paper is to create an automated system to evaluate and score a candidate's performance in the CEFR B2 English speaking test, removing the need for a human assessor and ensuring relevance in both global and Indian contexts. The framework of CEFR B2 Speaking Assessment and the automated scoring system is given in Figure~\ref{fig:blockdiagram}. Secondary aims involve developing an automated system capable of identifying vocabulary and proficiency levels. Furthermore, this model should generate CEFR B2 vocabulary and sentences. We introduced \textit{EvalYaks}, a set of six unique models, with four models dedicated to the primary aim and two to the secondary. \textit{EvalYaks} is trained by instruction tuning Mistral Instruct 7B v0.2 using the Low Rank Adaptation (LoRA), a parameter efficient fine tuning (PEFT) method.

Due to the scarcity of CEFR B2 English speaking assessment data specific to India, we generated a data set of simulated candidate responses and their evaluation scores using GPT-4. The data set was then refined with expert human feedback.

As a baseline, we first investigated the ability of leading LLMs, which are highly ranked on the LMSYS leaderboard \citep{lmsys}, to directly evaluate candidate responses by leveraging their intrinsic knowledge and prompt engineering. In addition, we examined the ability of LLMs to comprehend sentence structures and link vocabulary with CEFR proficiency levels. Our \textit{EvalYaks} suite of instruction fine-tuned (LoRA) models demonstrated superior performance in these tasks compared to standard LLMs. In our experiments, for each scenario, we used two sets of evaluation instructions for the LLM: one with contextual information and one without. Standard LLMs are not capable of satisfactory evaluation. Our findings indicate that a 7B parameter LLM, when instruction-tuned with specific, high-quality CEFR-aligned assessment data, can be employed for automated evaluation and scoring of CEFR B2 English speaking assessments.

\section{Cambridge's B2 First Speaking Assessment Framework}


We created \textit{EvalYaks} utilizing the Cambridge framework for the B2 First Exam. This speaking assessment is a component of the comprehensive exam that evaluates all aspects of language skills, including listening, reading, writing, and speaking. The primary objective of this exam is to assess the English language proficiency of the candidate at the CEFR B2 level. This framework is illustrated in Figure~\ref{fig:blockdiagram} and is explained in detail in the following subsections.

\subsection{Exam format}

The speaking test involves two candidates and two examiners. The first examiner acts as both an interlocutor and an assessor, facilitating the dialogue with the candidates. The second examiner serves as an assessor, monitoring the interaction.

The speaking test is divided into four sections \citep{cambridge_approach}. In the first section, the interlocutor individually assesses each candidate's social and interactional language skills. In the second section, each candidate speaks at length by comparing two photographs, followed by a brief reply from the second candidate. This section evaluates spoken production, assessing the candidate's ability to organize and express their thoughts coherently. In the third section, the candidates converse based on written prompts and instructions, followed by a collaborative decision-making task. This section assesses their pragmatic and strategic competencies as they exchange ideas, justify opinions, and use negotiation strategies. In the final section, the candidates engage in a discussion about the topics presented in the third section. This part assesses their capability to delve into the subjects and substantiate their viewpoints.

\begin{table}[h!]
\begin{singlespace}
\footnotesize
\centering
\caption{Cambridge B2 First speaking test's assessment scales excluding metrics for pronunciation \citep{cambridgeb2}.}
\begin{tabular}{>{\centering\arraybackslash}m{1cm} >{\arraybackslash}p{3.5cm} >{\arraybackslash}p{4cm} >{\arraybackslash}p{3.5cm}}
\toprule
\textbf{B2} & \textbf{Grammar and vocabulary} & \textbf{Discourse management} & \textbf{Interactive communication} \\
\midrule
\multirow{3}{*}{\centering \textbf{5}} & 
\parbox{3.5cm}{%
    \textbullet\ Shows a good degree of control of a range of simple and some complex grammatical forms.\\
    \textbullet\ Uses a range of appropriate vocabulary to give and exchange views on a wide range of familiar topics.
} & 
\parbox{4cm}{%
    \textbullet\ Produces extended stretches of language with very little hesitation.\\
    \textbullet\ Contributions are relevant and there is a clear organization of ideas.\\
    \textbullet\ Uses a range of cohesive devices and discourse markers.
} & 
\parbox{3.5cm}{%
    \textbullet\ Initiates and responds appropriately, linking contributions to those of other speakers.\\
    \textbullet\ Maintains and develops the interaction and negotiates towards an outcome.
} \\
\midrule
\textbf{4} & \multicolumn{3}{c}{\centering Performance shares features of Bands 3 and 5.} \\
\midrule
\multirow{3}{*}{\centering \textbf{3}} & 
\parbox{3.5cm}{%
    \textbullet\ Shows a good degree of control of simple grammatical forms, and attempts some complex grammatical forms.\\
    \textbullet\ Uses a range of appropriate vocabulary to give and exchange views on a range of familiar topics.
} & 
\parbox{4cm}{%
    \textbullet\ Produces extended stretches of language despite some hesitation.\\
    \textbullet\ Contributions are relevant and there is very little repetition.\\
    \textbullet\ Uses a range of cohesive devices.
} & 
\parbox{3.5cm}{%
    \textbullet\ Initiates and responds appropriately.\\
    \textbullet\ Maintains and develops the interaction and negotiates towards an outcome with very little support.
} \\
\midrule
\textbf{2} & \multicolumn{3}{c}{\centering Performance shares features of Bands 1 and 3.} \\
\midrule
\multirow{3}{*}{\centering \textbf{1}} & 
\parbox{3.5cm}{%
    \textbullet\ Shows a good degree of control of simple grammatical forms.\\
    \textbullet\ Uses a range of appropriate vocabulary when talking about everyday situations.
} & 
\parbox{4cm}{%
    \textbullet\ Produces responses which are extended beyond short phrases, despite hesitation.\\
    \textbullet\ Contributions are mostly relevant, despite some repetition.\\
    \textbullet\ Uses basic cohesive devices.
} & 
\parbox{3.5cm}{%
    \textbullet\ Initiates and responds appropriately.\\
    \textbullet\ Keeps the interaction going with very little prompting and support.
} \\
\bottomrule
\end{tabular}
\label{tab:language-assessment}
\end{singlespace}
\end{table}

\subsection{Candidate Assessment}

The assessor and the interlocutor assess the individual performances of the candidates based on the performance descriptors of the assessment scales \citep{ffrench2003development}. In the actual assessment, the assessor does not participate in the interaction and scores the criteria based on the descriptors given in Table~\ref{tab:language-assessment}. There are four criteria for assessment, viz., `grammar and vocabulary', `discourse management', `pronunciation', and `interactive communication'. The grammar part of the `grammar and vocabulary' criteria assesses the candidate's ability to apply grammatical rules and construct sentences utilizing a wide range of grammatical forms. Vocabulary focuses on the candidate's lexical range and ability to use appropriate vocabulary in different scenarios. `Discourse management' criteria assess the candidate's aptitude to produce organized and coherent discourse relevant to the questions/scenarios in monologues and interactions. Pronunciation focuses on the candidate's ability to produce easily comprehensible utterances. `Interactive communication' assesses the candidate's participation in a conversation, such as initiating and responding, taking turns without hesitation, and maintaining a conversation. 

In the present work, we focus on developing automated evaluators only for `grammar and vocabulary', `discourse management', and `interactive communication'. As such, Table~\ref{tab:language-assessment} lists the detailed assessment scales for Cambridge's B2 First speaking assessment excluding the pronunciation criteria.

\section{\textit{EvalYaks} Development Approach}

Our methodology for developing \textit{EvalYaks}, an automated system to assess English speaking proficiency of human test takers at the CEFR B2 level, consisted of four key stages. First, we generated simulated candidate conversation records, hereinafter referred to simply as conversation records that replicate realistic conversations corresponding to different parts of the assessment using the GPT-4 Turbo model (Jan'24). This data set contains conversations that mimicked actual assessment interactions between a candidate and an interlocutor or between a candidate, an interlocutor, and a partner, as required by different parts of the assessment, along with scores for different assessment criteria. The data were then carefully validated and aligned by experts.  We considered three assessment criteria: `grammar and vocabulary', `discourse management', and `interactive communication' (detailed in the rubric given in Table~\ref{tab:language-assessment}). Detailed information on this process can be found in Section~\ref{datagen}. Next, we examine the effectiveness of off-the-shelf cutting-edge language models in evaluating these conversation records by leveraging their built-in abilities. For this assessment, we used two types of prompts: one that included contextual data and another that did not. This stage informed us that instruction fine-tuning of models is necessary to achieve acceptable automated evaluation systems. The third stage involved the preparation of specialized instruction datasets to fine-tune a base language model to evaluate conversation records. For this, we created many more conversational records for various scenarios with the corresponding scores to train the model. Each conversation record was rigorously validated by experts. We used these conversation records and resources such as the English Vocabulary Profile (up to CEFR B2 level; \citep{englishprofile}) and the CEFR-SP WikiAuto dataset \citep{CEFR-SP} by adding specific instructions, making them suitable for instruction-tuning. See Section~\ref{instructdata_gen} for more details.  Finally, these validated data sets were utilized to create a suite of six models, one model each for detecting and generating CEFR B2 level vocabulary (\textit{EvalYaks} Vocab), and CEFR B2 sentence structures (\textit{EvalYaks} CEFR), and four models (\textit{EvalYaks} Parts 1-4) for automated evaluation of the performance of a candidate in the different parts of the CEFR B2 First English speaking assessment. This structured multistage approach ensures that our system is effective and reliable. Although we use the Cambridge CEFR B2 First exam as a reference, this approach can be extended to other English Qualifications.

\section{Dataset Generation and Evaluation Metrics}

Educational datasets frequently contain sensitive personal information, and for particular uses such as the CEFR B2 First English speaking evaluation, publicly available data are absent in the Indian context. Fortunately, synthetic data can be created using contemporary LLMs. We used OpenAI's GPT 4 Turbo (Jan '24) \citep{gpt4} to generate the synthetic dataset. This data set contains conversation records of the CEFR B2 English speaking assessment and the corresponding evaluations. The generation process is described below.

\subsection{CEFR B2 English speaking assessment conversation records generation and validation}\label{datagen}
We required a data set of the CEFR B2 English speaking assessment for Indian test takers. We developed prompts to generate the conversation records and employed the services of subject matter experts to validate and align the scores in the conversation records. 

\subsubsection{Prompt for synthetic data generation} 

A prompt acts as a directive for a language model to generate text that is not only pertinent, but also varied, realistic, and aligned with the target application \citep{prompt}. We needed to create conversations that would replicate various sections of the CEFR B2 First English speaking exam in an Indian and global context, along with scores for the three assessment criteria. This required the creation of a comprehensive and detailed prompt that included a wide range of information. The prompts for the four sections of the assessment were different since each part emphasizes different speaking skills, and the interactions differ significantly between the four parts. 

\begin{figure}[h!]
	\centering
		\includegraphics[width=\textwidth]{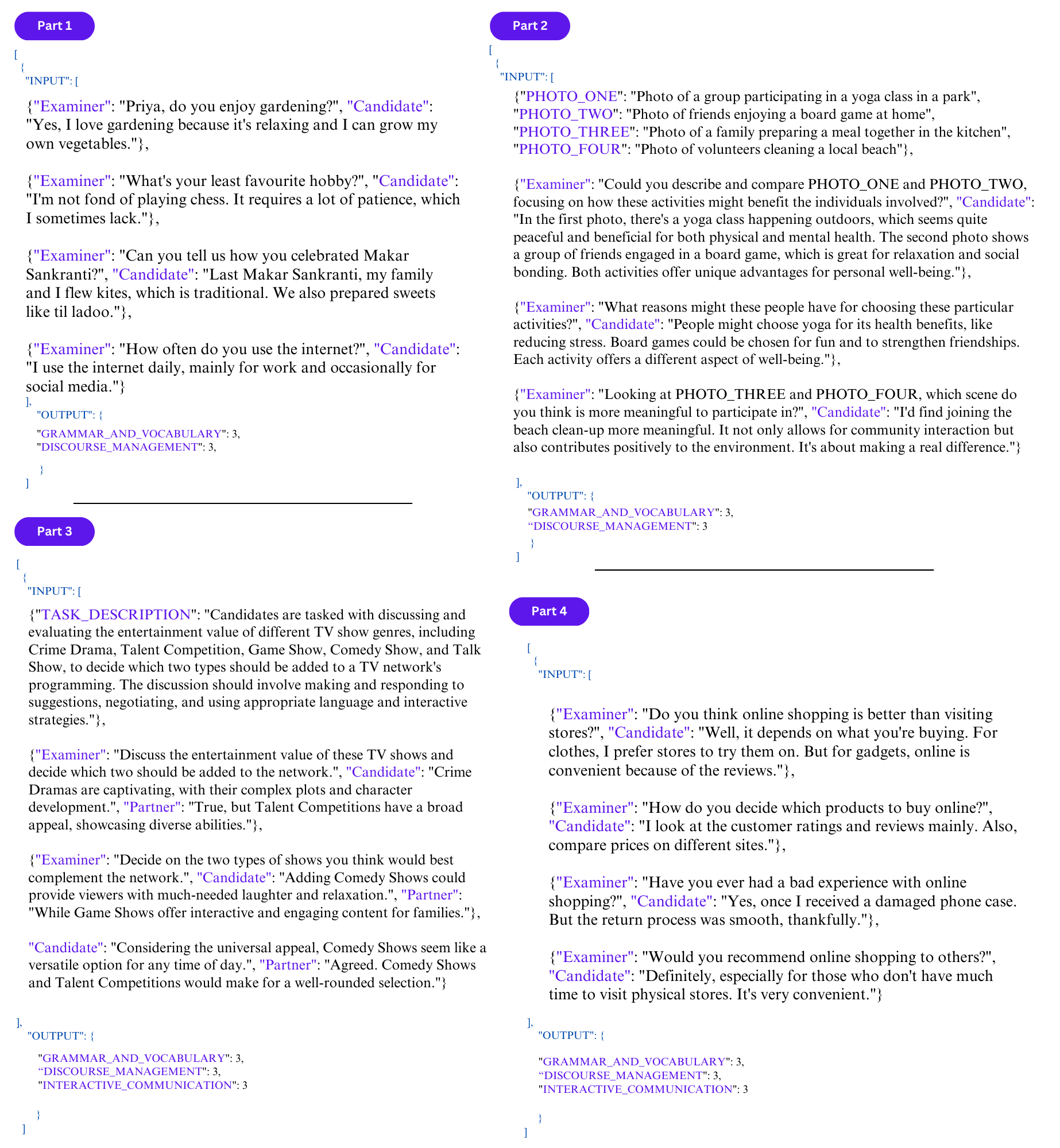}
	  \caption{Datapoint examples for different parts of the speaking assessment for instruction tuning.}\label{fig:data_format}
\end{figure}

Each prompt began with an overview of the particular part of the assessment. Following this, the model was instructed to produce conversations that corresponded to specific scores for each assessment criterion. For example, the output included assigning a score of 1 to both `grammar and vocabulary' and `discourse management'.  These directions used the Chain-of-Thought (CoT) prompting technique \citep{CoT}, which guided the model through the process of generating conversations corresponding to specific scores. After these directions, the output of the model consisting of conversations and scores was requested in JSON format. JSON's lightweight nature and ease of parsing make it particularly advantageous for production servers of LLMs, enhancing data interchange efficiency and reducing latency in API communications. Figure~\ref{fig:data_format} presents an example data set for each section of the assessment. The formats for the input and output data were tailored according to the unique objectives of each part of the CEFR B2 English speaking assessment.  

In addition, the prompt also contained context data that incorporate vocabulary up to the CEFR B2 level, assessment scales detailing the scoring for each criterion, and expectations were provided. Creating an automated evaluation system customized for the Indian context necessitates the model's ability to handle a wide range of India-specific information. Thus, India-specific context details such as Indian names, locations, festivals, professions, and common hobbies throughout the country were included. This information was used to craft conversations that blended Indian and global contexts. To ensure realistic data generation, a few-shot example strategy \citep{gpt3} was incorporated into the prompt utilizing example conversations from Cambridge's B2 First exam. Furthermore, the prompt contained examples of typical questions \citep{cambridgeb2} asked by the interlocutor during the speaking portion of the exam. \ref{appendix_generationprompt} has the prompts used to create the simulated CEFR B2 English speaking assessment conversation records.

\subsubsection{CEFR B2 English speaking assessment conversation records for parts 1-4}\label{datasetcombinations}

In parts 1, 2, and 4 of the CEFR B2 English speaking assessment, candidates have sequential dialogues with the interlocutor. Parts 1 and 2 collect only scores for `grammar and vocabulary' and `discourse management', as indicated in the example data shown in Figure~\ref{fig:data_format}. In part 3, candidates participate in a discussion among themselves and then interact with the interlocutor. Part 4 provides an opportunity for the candidates to further explore the topics of Part 3 with the interlocutor. For parts 3 and 4, the data include scores for `grammar and vocabulary', `discourse management', and `interactive communication'. 

In the data generation process, various conversations were formulated that aligned with different combinations of scores for `grammar and vocabulary', and `discourse management' for parts 1 and 2. Similarly, for parts 3 and 4, distinct conversations were generated that captured a variety of score combinations for `grammar and vocabulary', `discourse management', and `interactive communication'. The prompt instructions delineated the specific combinations to be generated with each API request to GPT-4 Turbo. The combinations were calibrated such that the score differences between the evaluation criteria within any part did not exceed two points, as differences in excess of two points are extremely rare in actual assessments. It is quite improbable that a candidate would score considerably lower in `grammar and vocabulary' (e.g., a score of one) while securing much higher scores in `discourse management' and `interactive communication' (e.g., scores of four or five).



We developed 25 user profiles to participate in the four-part CEFR B2 English speaking assessment for data generation. The profiles were grouped into five proficiency levels, each containing five profiles, with an average skill level ranging from 1 to 5. For parts 1 and 2, the groupings were determined by `grammar and vocabulary' and `discourse management'. For parts 3 and 4, `interactive communication' was also considered. The automated evaluation of the conversation records generated from the user profiles performed poorly with standard LLM when the ground truth scores were in the lower range (Section~\ref{results}). Consequently, we generated more low-scoring conversation records for training \textit{EvalYaks}. In total, we produced 3,060 data points (671, 667, 927, and 795 for parts 1-4, respectively). The data set for part 3 included a broader range of scenarios due to the addition of another speaker to the conversation, leading to more data points. Likewise, part 4 extended the variety of scenarios beyond those in parts 1 and 2.



\subsubsection{Data validation and alignment}

Given the intrinsic likelihood of errors and inconsistencies in text produced by large language models \citep{ji2023survey}, it is crucial to have subject matter experts validate the CEFR B2 English speaking assessment conversation and their corresponding scores in the conversation records. We carried out an alignment exercise to compare synthetically generated data with CEFR levels, enlisting human experts. Three English as a foreign language teachers, each with over four years of experience, assessed the quality of the generated data.

Each conversation record was subjected to separate evaluations by two out of three experts, allowing a dual perspective on its compliance with predetermined scores and assessment goals. The evaluations adhered to the criteria specified in Table~\ref{tab:language-assessment}, which outlines the expected outcomes for each score within those criteria. Each expert verified how well the conversation corresponded to the scores given for each criterion. When the conversation and the scores for each criterion aligned, the experts marked conversation record as correctly matched. In cases where discrepancies were identified, experts recorded the specific criteria where mismatches occurred and detailed the nature of these misalignments. Additionally, they suggested necessary modifications to the conversation to ensure alignment with the scores. These remarks were essential for identifying the required improvements for each conversation record.

Following their individual evaluations, the experts convened to discuss their findings. This session was pivotal in combining their insights and obtaining a consensus on the alignment of each data point with the intended ratings. After reviewing the comments collectively and engaging in an in-depth discussion, the experts agreed on the necessary adjustments to the conversations to better align them with the assessment scales. These final revisions were made to ensure that each conversation record accurately reflected the intended score for each criterion as presented in Table~\ref{tab:language-assessment}. This collaborative review and revision process was crucial for improving the overall accuracy and relevance of the data set in relation to the assessment objectives.



\subsection{Instruction dataset creation for fine-tuning \textit{EvalYaks}}\label{instructdata_gen}


\textit{EvalYaks} was trained utilizing the aligned conversation records to instruction tune a base language model. In instruction tuning, input-output pairs are enhanced with explicit instructions \citep{sanh2021multitask, naturalinstructions, supernaturalinstructions}. For our training process, we manually prepared three sets of instructions with varied information \citep{lima, alpaca} and also incorporated paraphrased versions of these instructions to generate additional training samples. The content of each instruction includes: \textit{(i)} The LLM's designated role as an evaluator of the CEFR B2 English speaking assessment and the detailed evaluation steps; \textit{(ii)} The LLM's designated role as an evaluator of the CEFR B2 English speaking assessment, the detailed evaluation steps, and performance descriptors; and \textit{(iii)} Evaluation steps and an example of output format.

We merged these instructions with the conversation records, incorporating special tokens to decipher various elements of the instruction, input, and output, to create instruction data points. We generate 1151, 1266, 2843, and 2085 instruction data points for parts 1 through 4 to train \textit{EvalYaks}. The three groups of instruction data point templates employed for training are detailed in \ref{appendix_training}. 

For instruction tuning \textit{EvalYaks} Vocab and CEFR models, we created instruction data for the English Vocabulary Profile (up to CEFR B2) and CEFR-SP WikiAuto datasets. The English Vocabulary Profile includes words, phrases, idioms, and collocations with different annotations of the appropriate CEFR levels from Cambridge. We used two sets of instructions (and paraphrased versions): one to identify and the other to generate words for specific CEFR levels, resulting in 3072 instruction data points from 5107 unique words. CEFR-SP includes 17k English sentences annotated with CEFR levels. From this, we used 7453 data points from the WikiAuto dataset, employing two sets of instructions (similar to the English Vocabulary Profile): identifying and generating sentences for specific CEFR levels, resulting in 19,142 instruction data points.


\subsection{Performance metrics}

To assess the performance of \textit{EvalYaks} suite of models and other standard LLMs to perform automated CEFR B2 English speaking assessment evaluation, we classified the results into four distinct levels: accurate, partly accurate, acceptable, and inaccurate. A response from the model is considered accurate when the scores across all assessment criteria are perfectly aligned with the reference scores. A response is classified as partly accurate if it matches the reference score on at least one assessment criterion. The classification of a response as acceptable involves a more nuanced criterion: It is labeled acceptable if the scores deviate by a margin of one, in the same direction from the reference scores, across any or all of the assessment criteria. For example, if the reference scores are (3,3), then the response scores of (2,2), (4,4), (2,3), (3,2), (3,4) and (4,3) are considered acceptable. However, combinations such as (2,4) or (4,2), despite both scores being within the acceptable range, are not considered acceptable due to the inconsistent deviation between the criteria. If any of the above conditions are not satisfied, then the response is considered to be inaccurate, including invalid responses. 

The acceptable accuracy of the model is calculated using the following formula with $N_{accu}$, $N_{part}$, $N_{acce}$ and $N$ being the number of accurate, partly accurate, acceptable, and total responses generated.
\begin{equation}
\centering
    acceptable\_accuracy = \frac{N_{accu} + N_{part} + N_{acce}}{N}
\end{equation}

An average acceptable accuracy is defined taking into account all four parts of the CEFR B2 English speaking assessment as
\begin{equation}
\centering
average\_acceptable\_accuracy = \frac{\sum_{i=1}^{4}{acceptable\_accuracy}_i}{4}\,,
\end{equation}

where $acceptable\_accuracy_i $ is the acceptable accuracy for each part, with \( i =1,2,3,4\) referring to the different parts of the assessment.

In addition to acceptable accuracy, we also calculated the degree of variation (DOV) in the responses. DOV refers to the average variation of the assessment scores relative to its reference value. DOV in the response for parts one and two is defined as
\begin{equation} \label{DOV12}
\centering
    DOV_{1} = DOV_{2} = \sum_{i=1}^N\frac{\mid GV_{ri}  - GV_{ai}\mid + \mid DM_{ri} - DM_{ai} \mid}{2N}\,,
\end{equation}

and for parts three and four is defined as
\begin{equation} \label{DOV34}
\centering
    DOV_{3}=DOV_{4}=\sum_{i=1}^N\frac{\mid GV_{ri}-GV_{ai}\mid+\mid DM_{ri}-DM_{ai} \mid + \mid IC_{ri} - IC_{ai} \mid}{3N}\,,
\end{equation}
where GV stands for `grammar and vocabulary', DM for `discourse management', and IC for `interactive communication'. The subscripts `r' and `a' are used to differentiate between the references and the actual scores. For DOV, smaller is better.

\section{Training \textit{EvalYaks}}

Each of the six models that are part of the \textit{EvalYaks} suite was trained separately with mistral-7b-instruct-v0.2 as the base model on which Low Rank Adaptation (LoRA) \citep{lora} was used for parameter and memory efficient instruction fine-tuning. By introducing two smaller matrices for weight updates via low-rank decomposition, LoRA enables adaptations to new data without altering the original weight matrix, which remains unchanged. These LoRA adapters can be merged with the base model during inference. During training, individual adapters were developed for each dataset, such as one adapter specifically for automated evaluation and scoring of CEFR B2 speaking assessment Part 1, another for part 2, and so on. We experimented with multiple dataset combinations, including aggregating parts 1 and 2, parts 3 and 4, all four parts together, and all parts in conjunction with the English Vocabulary Profile and CEFR-SP WikiAuto data sets. Upon evaluation, it was observed that the models equipped with dedicated adapters for each individual data set exhibited superior performance.

We experimented with combinations of hyper-parameters that control the rank of the update matrices $r$ and the LoRA scaling factor $\alpha$, such as ($r,\alpha$) = (64, 16), (256, 128), and (256, 512). The best adapters were obtained with $r=256$ and $\alpha=128$ with a lora dropout of 0.1. In our instruction tuning process, we trained for 5 epochs using a cosine learning rate schedule starting at $2e^{-4}$, employing the AdamW optimizer \citep{adamw} with a weight decay of 0.001. The adapters were trained using bfloat16 precision on 1 or 2 NVIDIA A100 GPUs (each with 80 GB of VRAM), 3 or 4 RTX A6000 Ada GPUs (each with 48 GB of VRAM), or 3 or 4 RTX A6000 GPUs (each with 48 GB of VRAM). This training was conducted on servers available through RunPod and on local servers. 

Thus, `\textit{EvalYaks}' refers to a family of six distinct models, each resulting from the integration of an individual adapter with the base model with the best performing hyperparameters. 


\section{Results and Discussion}\label{results}

We investigated the performance of \textit{EvalYaks} and standard LLMs without LoRA in the automated assessment of conversation records of the CEFR B2 English speaking assessment. The procedure began with the evaluation of LLMs using prompts both without and with performance descriptors. We conducted an extensive analysis of the \textit{EvalYaks} Vocab and CEFR models, as well as \textit{EvalYaks} Part 1-4 models.

The assessment was conducted by providing the LLMs with two different sets of prompts, which included comprehensive instructions and conversations from the test data. The first prompt specified the LLM's role as a CEFR B2 English speaking assessment evaluator, detailed the evaluation steps, and included the conversation to be assessed. The second prompt included the performance descriptors from Table \ref{tab:language-assessment} along with the content from the first prompt. These descriptors served as a reference for the model to rate the assessment criteria. Both sets of prompts used in the evaluation are presented in \ref{appendix_evaluation}.



\subsection{Performance Analysis of standard LLMs without LoRA}\label{LLMs_explain}

We first investigated the intrinsic abilities of leading (as of March 2024) LLMs to perform the automated CEFR B2 English speaking evaluation task. We utilized 11 different LLMs (both proprietary and open-source of different sizes) that are at the top of the LMSYS leaderboard \citep{lmsys}. These include Gemma 7B \citep{gemma}, Mistral Instruct v0.2 \citep{mistral}, Llama2 7B Chat \citep{llama2}, Vicuna 33B \citep{vicuna}, Mixtral Instruct v0.1 \citep{mixtral}, Llama2 70B Chat \citep{llama2}, Qwen 72B Chat \citep{qwen}, GPT 3.5 (Jan '24), Claude Haiku (Mar '24) \citep{claude}, Gemini Pro 1.0 \citep{gemini}, and Mistral Medium. These models vary in complexity, ranging from 7 billion parameters to some claiming to have hundreds of billion parameters. 

\begin{table}[h!]
\begin{singlespace}
\footnotesize
\centering
\caption{The average acceptable accuracy of state-of-the-art LLMs without LoRA for CEFR B2 English speaking assessment using prompts without and with performance descriptors.}\label{combined_llms}
\begin{tabular*}{\textwidth}{@{\extracolsep{-0.1cm}}cccc}
\toprule
Model &  Organization & Without performance  & With performance \\ 
& & descriptors & descriptors \\
\midrule
Gemini Pro 1.0 & Google DeepMind & \textbf{69\%} & \textbf{82\%}\\
Vicuna 33B & LMSYS & 66\% & 66\%\\
Claude Haiku (Mar '24) & Anthropic & 62\% & 61\%\\
Llama2 70B Chat & Meta & 56\% & 55\%\\
Llama2 7B Chat & Meta & 53\% & 64\%\\
Mixtral Instruct v0.1 & Mistral & 47\% & 53\%\\
Mistral Medium & Mistral & 46\% & 80\%\\
Qwen 72B Chat & Alibaba & 46\% & 74\%\\
GPT 3.5 (Jan '24) & OpenAI & 43\% & 62\%\\
Mistral Instruct v0.2 & Mistral & 40\% & 47\%\\
Gemma 7B & Google DeepMind & 29\% & 57\%\\
\bottomrule
\end{tabular*}
\end{singlespace}
\end{table}


Table~\ref{combined_llms} presents the average acceptable accuracy metric achieved by all evaluated models when prompted with and without performance descriptors. It is observed that Gemini Pro 1.0 outperforms the other models, achieving an average acceptable accuracy of 69\% without performance descriptors and 82\% with them. Providing contextual information, such as details about the assessment scales, clearly enhances the performance of Gemini Pro 1.0. This trend is consistent across most of the other LLMs, except for Claude haiku (Mar'24) and Llama2 70B Chat, which saw a slight decrease of 1\% in their average acceptable accuracy, and Vicuna 33B, which showed no change. Meanwhile, the Gemma 7B, Qwen 72B Chat, and Mistral Medium models demonstrated substantial performance improvements, with increases of 28\%, 28\%, and 34\%, respectively, in their average acceptable accuracy.


It is notable that the model's size does not directly correlate with its capacity to assess learners' performance in the CEFR B2 English speaking evaluation. This implies that the size of the model does not inherently imply its proficiency for a particular task. However, adding contextual information to the prompt can greatly enhance the LLMs' ability to evaluate learners' performance in the CEFR B2 English speaking assessment.



\subsection{Performance analysis of \textit{EvalYaks} Vocab and CEFR models}

We explored the ability of a relatively small, instruction-tuned language model to grasp the inherent structure of sentences and link vocabulary to CEFR proficiency levels through instruction tuning. The \textit{EvalYaks} Vocab model, trained on a dataset created from the English Vocabulary Profile, achieves an acceptable accuracy of 100\%. The model's accurate response rate is 85\%, while its acceptable response rate is 15\%. This suggests that the model can reliably associate vocabulary with the appropriate CEFR levels through instruction tuning.

\textit{EvalYaks} CEFR model, trained on an instructional dataset derived from the CEFR-SP WikiAuto dataset, achieved a satisfactory accuracy rate of 96.25\% for determining the CEFR level of sentences or generating sentences to align with specific CEFR levels. The model produced accurate responses 65\% of the time, acceptable responses 31.25\% of the time, and inaccurate responses 3.75\% of the time. This indicates that the model is capable of evaluating sentence-level CEFR alignment, with most of its responses being either accurate or acceptable.

\subsection{Performance analysis of \textit{EvalYaks} Part 1-4 models}

\begin{figure}[h!]
	\centering
		\includegraphics[width=\textwidth]{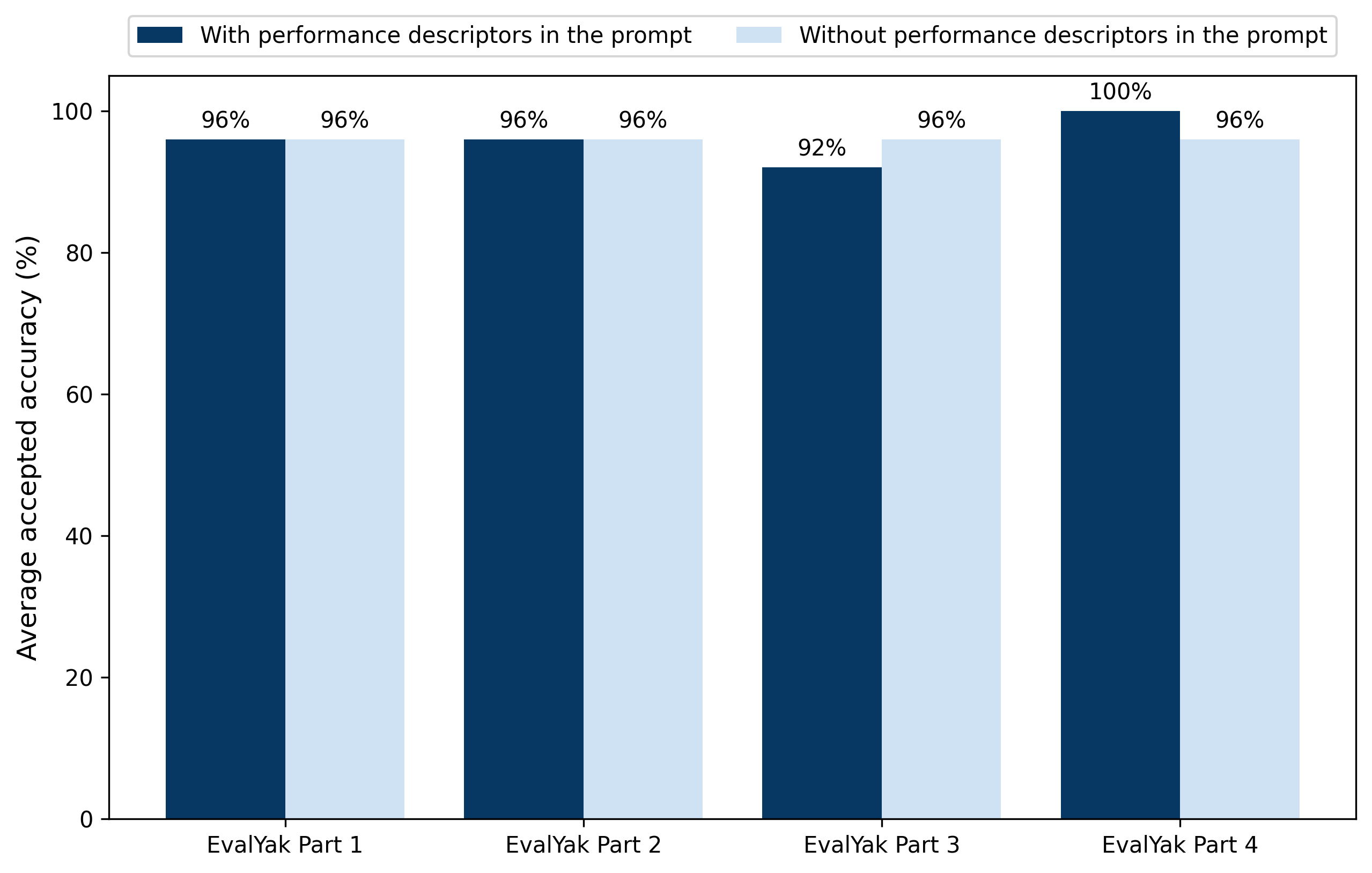}
	  \caption{The average acceptable accuracy of \textit{EvalYaks} part 1-4 models in CEFR B2 English speaking assessment with and without performance descriptors in the prompt.}\label{fig:adapters}
\end{figure}

\begin{figure}[h!]
	\centering
		\includegraphics[width=\textwidth]{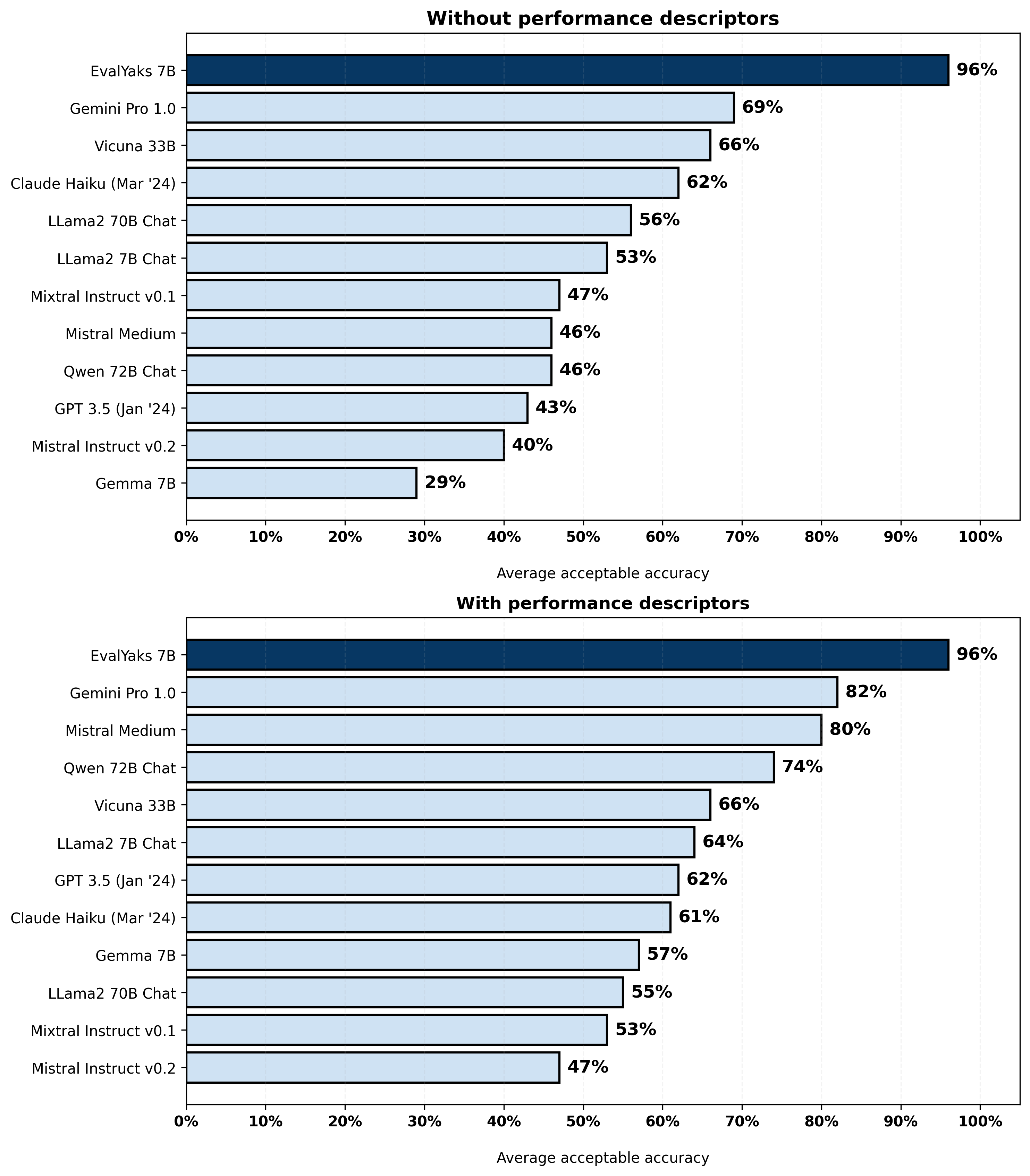}
	  \caption{The average acceptable accuracy of state-of-the-art LLMs without LoRA in comparison with \textit{EvalYaks} part 1-4 models in CEFR B2 English speaking assessment with and without performance descriptors in the prompt.}\label{fig:average_acceptable}
\end{figure}

\begin{figure}[h!]
	\centering
		\includegraphics[width=\textwidth]{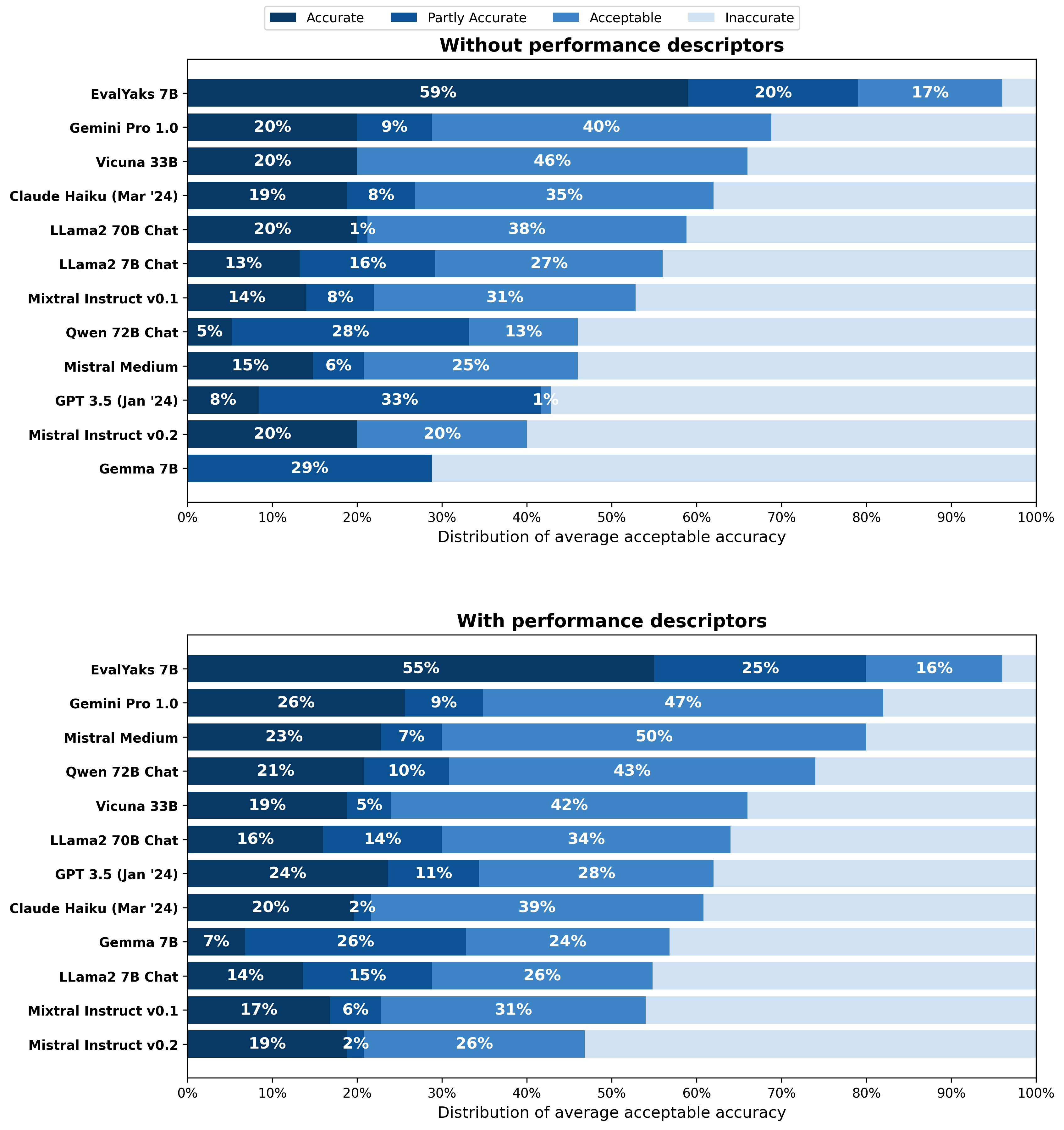}
	  \caption{The distribution of average acceptable accuracy of state-of-the-art LLMs without LoRA in comparison with \textit{EvalYaks} part 1-4 models in CEFR B2 English speaking assessment with and without performance descriptors in the prompt.}\label{fig:average_acceptable_stacked}
\end{figure}

We performed an in-depth analysis to understand whether instruction tuning with high-quality CEFR-aligned conversational records could improve the performance of a 7B parameter model compared to other leading LLMs without fine-tuning.  Figure~\ref{fig:adapters} presents the average acceptable accuracy of the \textit{EvalYaks} models in different parts of the CEFR B2 English speaking assessment, both with and without the inclusion of performance descriptors in the prompts.  The \textit{EvalYaks} models maintain a notably high average acceptable accuracy, with 96\% in part one when prompted with and without performance descriptors. This level of accuracy is consistent with that in part 2. In part 3, the average acceptable accuracy of the model 92\% and 96\% when prompted with and without performance descriptors, respectively. In the case of part 4, the model achieves an average acceptable accuracy of 100\% with performance descriptors in the prompt and 96\% without performance descriptors in the prompt. 

\textit{EvalYaks} models, tuned with high-quality CEFR-aligned conversation records, demonstrated an average acceptable accuracy of 96\% whether performance descriptors were included or not, as illustrated in Figure~\ref{fig:average_acceptable}. In contrast, the next best-performing model, Gemini Pro 1.0, showed an average acceptable accuracy of 69\% for prompts without performance descriptors and 82\% for prompts with performance descriptors. The \textit{EvalYaks} model family comprises 7B parameter models, while it is speculated that Gemini Pro 1.0 has hundreds of billions of parameters. The base model for \textit{EvalYaks}, Mistral Instruct v0.2, achieved an acceptable accuracy rate of 40\% for prompts without performance descriptors and 47\% for those with performance descriptors, which is significantly lower than the average acceptable accuracy of \textit{EvalYaks} models.

We examined the models' performance by looking at the distribution of accurate, partly accurate, acceptable, and inaccurate responses averaged across the four components of the speaking assessment. Figure~\ref{fig:average_acceptable_stacked} shows that the \textit{EvalYaks} models possess the highest percentage of accurate responses for the prompts with and without performance descriptors. The accurate response rates are 59\% for the evaluation with the prompts lacking performance descriptors and 55\% for the evaluation using the prompts with performance descriptors. In particular, Gemini Pro 1.0, which ranked second in performance, achieved only 20\% and 26\% accurate responses in the evaluations using the prompts without and with performance descriptors, respectively. The increased average acceptable accuracy in other LLMs can be attributed to an increased number of acceptable responses rather than accurate or partly accurate ones.

\subsubsection{Analysis of \textit{EvalYaks} Part 1-4 models without performance descriptors}

\begin{figure}[h!]
	\centering
		\includegraphics[width=\textwidth]{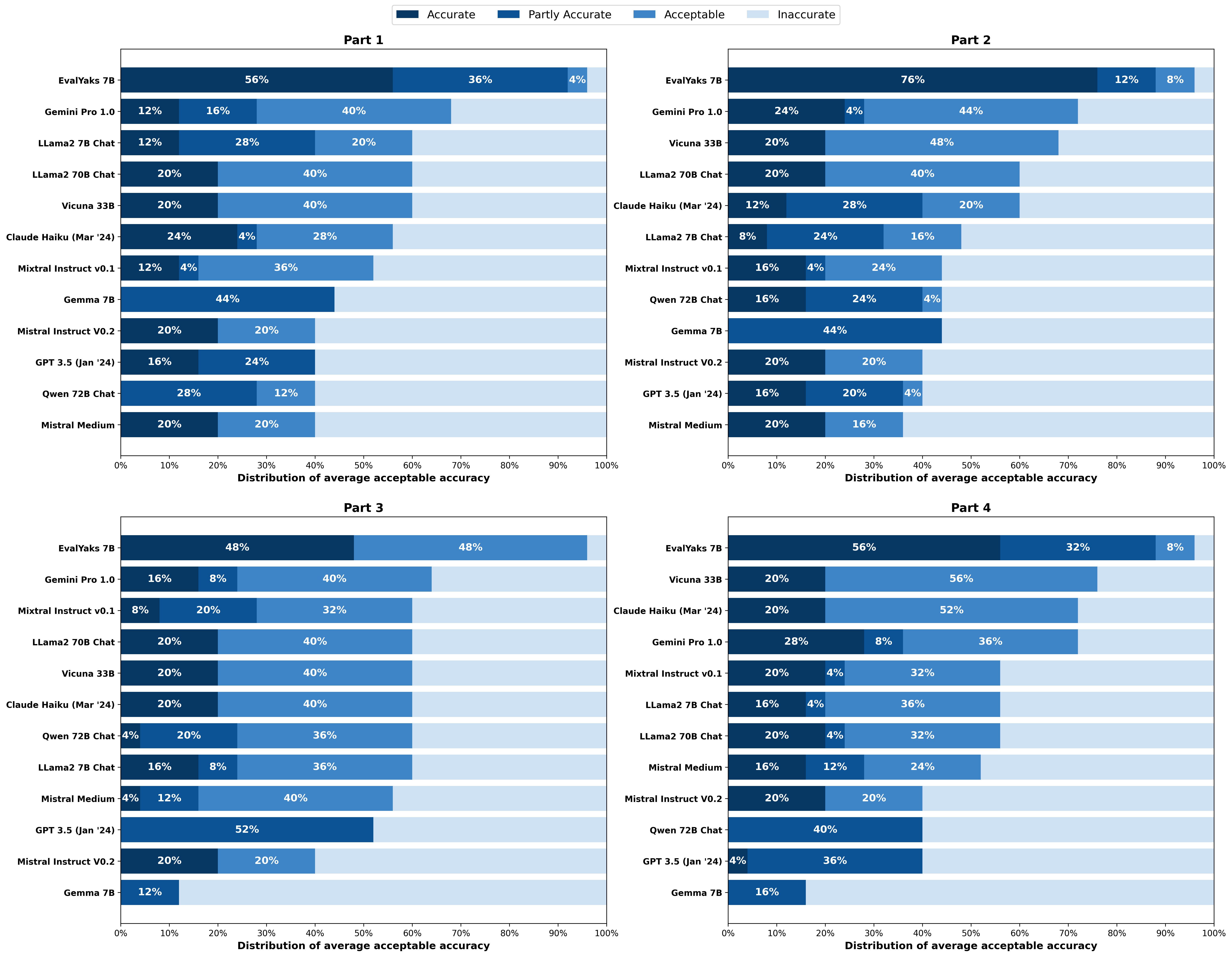}
	  \caption{The distribution of acceptable accuracy of state-of-the-art LLMs without LoRA in comparison with \textit{EvalYaks} part 1-4 models using prompts without performance descriptors.}\label{fig:stacked_graph_without_performancedesc}
\end{figure}

Figure~\ref{fig:stacked_graph_without_performancedesc} illustrates the models' performance in terms of the distribution of accurate, partly accurate, acceptable, and inaccurate responses in different parts of the evaluation without performance descriptors. The figure clearly shows that \textit{EvalYaks} models outperform others in all sections. \textit{EvalYaks} part 1-4 models exhibit the highest proportion of accurate responses, with percentages of 56\%, 76\%, 48\%, and 56\% for parts 1 to 4. In each part of the assessment, the percentage of accurate responses exceeds that of partly accurate and acceptable responses, except in part 3, where the percentages of accurate and acceptable responses are identical. 

The analysis revealed that while standard LLMs achieved a reasonable level of acceptable responses, their accurate and partly accurate responses were often lower. This consistent performance underscores the effectiveness of targeted instruction tuning in enhancing the accuracy and reliability of automated evaluators for speaking assessments.

\subsubsection{Analysis of \textit{EvalYaks} Part 1-4 models with performance descriptors}

\begin{figure}[h!]
	\centering
		\includegraphics[width=\textwidth]{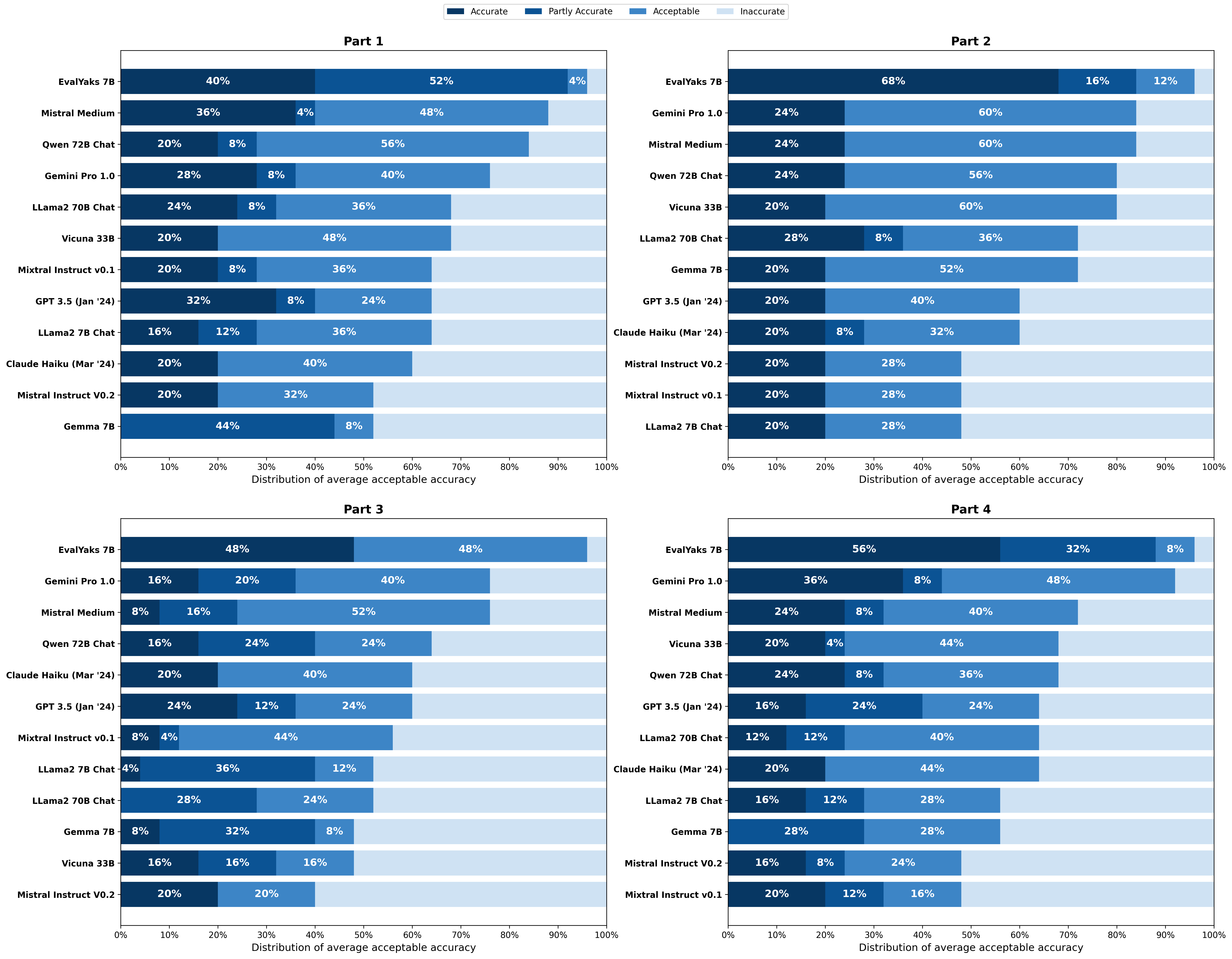}
	  \caption{The distribution of acceptable accuracy of state-of-the-art LLMs without LoRA in comparison with \textit{EvalYaks} part 1-4 models using prompts with performance descriptors.}\label{fig:stacked_graph_with_performancedesc}
\end{figure}

Figure~\ref{fig:stacked_graph_with_performancedesc} demonstrates the model's performance in the distribution of accurate, partly accurate, acceptable, and inaccurate responses in various parts of the CEFR B2 English speaking exam when using performance descriptors. Notably, \textit{EvalYaks} achieves the highest accuracy rate in all parts of the assessment, with percentages of 40\%, 68\%, 48\%, and 56\% for parts one, two, three, and four, respectively. Apart from part three, \textit{EvalYaks} models also show a higher proportion of partly accurate responses. Although \textit{EvalYaks} maintain a consistent average acceptable accuracy, their performance is reduced compared to evaluations without performance descriptors. Other models, which were not trained with CEFR-aligned conversation records, generally exhibit a higher rate of acceptable responses compared to accurate and partly accurate responses, with the exception of Gemma 7B and Llama2 7B in part three.

The models' average acceptable accuracy fluctuated across various parts of the speaking assessment. In part 1, Mistral Medium ranks second with an average acceptable accuracy of 88\%. Similarly, Gemini Pro 1.0 maintains the second position in parts 2, 3, and 4, achieving average acceptable accuracies of 84\%, 76\%, and 94\%, respectively. As previously mentioned, despite the high overall average accuracy of the models, it is important to recognize that in part 1, 48\% of responses were acceptable, and in parts 2, 3, and 4, the acceptable responses were 60\%, 40\%, and 48\%, respectively. Upon fine-tuning with CEFR-aligned conversation records, the model that initially scored the lowest or second lowest in different parts of the assessment exhibited the most substantial improvement, eventually achieving the best performance in all components of the speaking assessment. This underscores the effectiveness of specific instruction tuning in enhancing model performance for designated tasks. 

\subsubsection{Degree of variation in \textit{EvalYaks} part 1-4  models}

The second performance metric used to analyze the models is the degree of variation (DOV) in the scores assigned by the model for the input conversation in different parts of the assessment. The equations governing DOV in scores are given in Equations~\ref{DOV12} and \ref{DOV34}. The DOV obtained for different parts of the assessment when prompted without and with the performance descriptors are given in Table~\ref{llms_dov}. In the context of DOV, a lower value indicates better performance, meaning the model's outputs are closer to the expected standards or answers. The \textit{EvalYaks} models perform much better than the other models with a DOV of 0.34, 0.20, 0.52, 0.31, and 0.34 in parts 1, 2, 3, and 4 and overall, respectively when prompted without performance descriptors. During evaluations that incorporate performance descriptors within the prompts, the \textit{EvalYaks} models demonstrate consistently low DOV values, recording scores of 0.34, 0.26, 0.52, 0.31, and 0.36 for parts 1, 2, 3, and 4, and overall, respectively. The \textit{EvalYaks} models perform the best in part 2 and perform the their worse in part 3 regardless of the prompt type. Interestingly, the performance of the model in part 2 slightly deteriorates when prompts included descriptors, giving a higher overall DOV of 0.36 compared to 0.34 in the evaluation using the prompts without descriptors.

\begin{table}[h!]
\begin{singlespace}
\footnotesize
\centering
\caption{The degree of variation (DOV) in the scores across parts 1-4 of the assessment for prompts without and with performance descriptors (lower values are more desirable).}\label{llms_dov}
\begin{tabular*}{\textwidth}{@{\extracolsep{\fill}}lccccc}
\toprule
\multicolumn{6}{c}{Without performance descriptors} \\
\cmidrule{1-6}
Model & Part 1 & Part 2 & Part 3 & Part 4 & Overall \\
\midrule 
\textit{EvalYaks} & \textbf{0.34} & \textbf{0.20} & \textbf{0.52} & \textbf{0.31} & \textbf{0.34} \\
Gemini Pro 1.0 & 1.06 & 1.04 & 1.01 & 0.89 & 1.00 \\
Vicuna 33B & 1.32 & 1.08 & 1.29 & 1.08 & 1.19 \\
Claude haiku (Mar'24) & 1.18 & 1.26 & 1.36 & 1.08 & 1.22 \\
Qwen 72B Chat & 1.22 & 1.08 & 1.37 & 1.16 & 1.21 \\
Mixtral Instruct v0.1 & 1.13 & 1.35 & 1.39 & 1.15 & 1.26 \\
Llama2 70B Chat & 1.25 & 1.28 & 1.39 & 1.33 & 1.31 \\
GPT 3.5 (Jan'24) & 1.22 & 1.44 & 1.35 & 1.27 & 1.32 \\
Gemma 7B & 1.26 & 1.46 & 1.51 & 1.43 & 1.36 \\
Mistral Medium & 1.42 & 1.37 & 1.44 & 1.23 & 1.36 \\
Llama2 7B Chat & 1.32 & 1.56 & 1.39 & 1.33 & 1.40 \\
Mistral Instruct v0.2 & 1.65 & 1.72 & 1.99 & 1.79 & 1.79 \\
\midrule
\multicolumn{6}{c}{With performance descriptors} \\
\cmidrule{1-6}
Model & Part 1 & Part 2 & Part 3 & Part 4 & Overall \\
\midrule 
\textit{EvalYaks} & \textbf{0.34} & \textbf{0.26} & \textbf{0.52} & \textbf{0.31} & \textbf{0.36} \\
Gemini Pro 1.0 & 1.06 & 1.04 & 0.95 & 0.64 & 0.92 \\
Vicuna 33B & 1.04 & 0.94 & 1.17 & 1.11 & 1.07 \\
Claude haiku (Mar'24) & 1.24 & 1.16 & 1.37 & 1.13 & 1.23 \\
Qwen 72B Chat & 1.22 & 1.08 & 1.01 & 0.93 & 1.06 \\
Mixtral Instruct v0.1 & 1.10 & 1.42 & 1.52 & 1.23 & 1.32 \\
Llama2 70B Chat & 0.92 & 0.88 & 1.29 & 1.01 & 1.03 \\
GPT 3.5 (Jan'24) & 0.88 & 1.02 & 0.96 & 0.93 & 0.99 \\
Gemma 7B & 1.26 & 1.02 & 1.28 & 1.17 & 1.18 \\
Mistral Medium & 0.78 & 0.92 & 1.01 & 0.88 & 0.90 \\
Llama2 7B Chat & 1.18 & 1.36 & 1.25 & 1.15 & 1.24 \\
Mistral Instruct v0.2 & 1.30 & 1.34 & 1.89 & 1.37 & 1.48 \\
\bottomrule
\end{tabular*}
\end{singlespace}
\end{table}

In contrast to the \textit{EvalYaks}, which exhibits a slight increase in DOV when evaluated with prompts containing performance descriptors, most other LLMs (except Claude haiku (Mar’24) and Mixtral Instruct v0.1), which did not undergo instruction tuning with CEFR-aligned conversation records, show improved performance under the same conditions. The model ranking next in performance, Gemini Pro 1.0, demonstrated an overall DOV of 1 when evaluated without performance descriptors in the prompts. During the evaluation with prompts incorporating performance descriptors, Mistral Medium emerged as the second-best performing model, achieving a DOV of 0.90. The difference in the DOV between the \textit{EvalYaks} models and the second-best performing models is 0.66 for evaluation using prompts without performance descriptors and 0.54 for those with performance descriptors.

Mistral Instruct v0.2 emerges as the model with the lowest performance, recording DOV scores of 1.65, 1.72, 1.99, 1.79, and 1.79 when assessed with prompts without performance descriptors for parts 1, 2, 3, and 4 and overall, respectively. Conversely, when evaluated with prompts containing descriptors, it shows slightly improved DOV scores of 1.30, 1.34, 1.89, 1.37, and 1.48 for parts 1, 2, 3, and 4 and overall, respectively. Despite its relatively lower performance, it is critical to note that the \textit{EvalYaks} models, which were instruction tuned with CEFR-aligned conversation using Mistral Instruct v0.2 as the base model, displayed superior performance compared to much larger models. This underscores the effectiveness of targeted instruction tuning in improving the model's performance for specific use cases. 

\section{Conclusion}


In this study, we created \textit{EvalYaks}, a collection of six models aimed at the automated assessment of CEFR B2 English speaking assessment parts 1 through 4, as well as CEFR Vocabulary and sentence structures. These models were trained by instruction fine-tuning the base Mistral Instruct V0.2 with a high-quality, well-aligned candidate conversation records dataset that we developed. The performance of the Automated Evaluators was evaluated based on the average acceptable accuracy and Degree of variation (DOV) in the scores.

\textit{EvalYaks} surpassed its primary competitor, Gemini Pro 1.0, by threefold in terms of accurate score predictions and doubled its effectiveness when performance descriptors were included. The base model of \textit{EvalYaks}, Mistral Instruct V0.2, ranked near the lowest among 11 models in the CEFR B2 English assessment. \textit{EvalYaks} consistently demonstrated strong performance in this assessment, registering a DOV of 0.34 and 0.36 for scenarios without and with performance descriptors, respectively. This was roughly one-third of the variation seen in Gemini Pro 1.0, which had a DOV of 0.66 more than \textit{EvalYaks}. The subsequent five top models demonstrated DOVs ranging from 0.19 to 0.31 more in contrast to Gemini Pro 1.0, indicating three to four times greater effectiveness of \textit{EvalYaks} among the other leading models. Overall, \textit{EvalYaks} showed consistent excellence across all evaluated metrics.

Our research confirms that large language models can effectively understand sentence structures and match vocabulary to CEFR levels. The alignment training for \textit{EvalYaks} models that focused on Cambridge B2 First is applicable to the other Cambridge English Qualifications. This adaptability suggests that similar models could be successfully used for automated CEFR speaking assessments, offering scalable and resource-efficient solutions.

Although LoRA adapters have shown potential, they may not suffice for a robust, production-level system on their own. To boost reliability, we plan to incorporate \textit{EvalYaks} with language agents \citep{wang2024survey}, Direct Preference Optimization (DPO), and reflection \citep{madaan2024self,shinn2024reflexion,rafailov2024direct} into the workflow. The reflective agents will enable self-improvement and evaluation of the scores during the iterations preceding the final outcomes. When \textit{EvalYaks} produce `partly accurate', `acceptable', and `inaccurate' responses, employing preference sampling with human experts can fine-tune these scores. We also plan to add descriptive feedback to offer deeper insights into the performance, highlighting their strengths and areas for improvement. This enhancement not only displays detailed scores but also elucidates the reasoning behind them, fostering learning through targeted feedback. Furthermore, adding memory to the \textit{EvalYaks} can connect assessment results with curated learning content, enriching the e-learning experience through a synergistic relationship between assessment and educational content. As \textit{EvalYaks} are language models, we have used text inputs for scoring. However, recognizing the crucial role of pronunciation in speaking assessments, a multi-modal system that can evaluate both textual and auditory elements of speech is a future area of study.

\section*{Acknowledgements}

This publication/presentation/research report has made use of the English Vocabulary Profile.  This resource is based on extensive research using the Cambridge Learner Corpus and is part of the English Profile programme, which aims to provide evidence about language use that helps to produce better language teaching materials. See \url{http://www.englishprofile.org} for more information. 







\appendix
\section{Data generation prompts}\label{appendix_generationprompt}

The prompt used with GPT 4 Turbo (Jan'24) to generate synthetic conversations and assessments is given below.

\vspace{0.2cm} 

\noindent \textit{In part one, candidates engage in a one-on-one conversation with the ``Examiner" covering personal details, routines, preferences, etc. This segment aims to evaluate their spontaneous communication skills in everyday situations. ``Candidate" should relax as the conversation starts, focusing on listening attentively and providing detailed answers, avoiding mere yes or no responses. While they should elaborate with reasons and examples, overly lengthy responses are not necessary at this stage.}

\vspace{0.2cm}

\noindent \textit{\textbf{Data generation instructions:}}

\begin{itemize}
    \item \textit{I want to create a conversational dataset for several combinations of scores in GRAMMAR AND VOCABULARY and DISCOURSE MANAGEMENT with the provided context. I need the output in the JSON format as given in the `Data Format'.}
    
     \item \textit{I first want to start with BAND 1 in GRAMMAR AND VOCABULARY \& BAND 1 in DISCOURSE MANAGEMENT.}

     \item \textit{Here the conversation is between the ``Examiner" and the ``Candidate". The conversation lasts for 4 exchanges (4 questions from Examiner and 4 responses from the Candidate). And this must be categorised as ``INPUT" in the JSON. }

     \item \textit{Refer to the `Typical Questions' and create similar questions for the conversation}

     \item \textit{Refer to `Indian Context' to create to generate Indian context specific questions and responses from time to time.}

     \item \textit{Use the A1, A2, B1 \& B2 CEFR vocabulary given in the context.}

     \item \textit{As the OUTPUT, I want you to provide a BAND score of 1 each to Key `GRAMMAR AND VOCABULARY' \& Key `DISCOURSE MANAGEMENT' after reading through the performance descriptors.}

     \item \textit{Please review the provided ``Sample Interactions" and their corresponding ratings to ensure the appropriate level of thoroughness in creating your responses.}

\end{itemize}

\noindent \textit{\textbf{Data Format:} \{data\_format\}} 

\vspace{0.2cm}

\noindent \textit{\textbf{Typical Questions:} \{typical\_questions\}} 

\vspace{0.2cm}

\noindent \textit{\textbf{Performance Descriptors:}}

\begin{itemize}
\item \textit{GRAMMAR AND VOCABULARY: \{performance descriptor grammar and vocabulary\}}
\item \textit{DISCOURSE MANAGEMENT: \{performance descriptor for discourse management\}}
\end{itemize}

\noindent \textbf{\textit{CEFR Vocabulary: }}

\begin{itemize}
    \item \textit{A1 Words: \{A1\_words\}}
    \item \textit{A2 Words: \{A2\_words\}}
    \item \textit{B1 Words: \{B1\_words\}}
    \item \textit{B2 Words: \{B2\_words\}}
\end{itemize}

\noindent \textit{\textbf{Indian Context:}}

\begin{itemize}
\item \textit{Names: \{indian\_names\}}
\item \textit{Places: \{indian\_places\}}
\item \textit{Festivals: \{indian\_festivals\}}
\item \textit{Professions: \{indian\_professions\}}
\item \textit{Hobbies: \{indian\_hobbies\}}
\end{itemize}

\noindent \textit{\textbf{Sample Interactions:} \{sample\_interactions\}} 

\section{Instruction datapoint template} \label{appendix_training}

The three sets of instruction datapoint for part one of the CEFR B2 English speaking assessment are given below. 

\subsection{Instruction datapoint without performance descriptors}

\vspace{0.2cm}

\textit{\textless s\textgreater [INST]}

\vspace{0.2cm}

\textit{\textbf{Role:}}
\begin{itemize}
    \item \textit{You are an assessor of the CEFR B2 English speaking assessment. You are an expert in this field with several years of experience.}
    \item \textit{You will be given a conversation between an Examiner and a Candidate, and your task is to give scores for two metrics for the responses given by the ``Candidate" in the conversation.} 
\end{itemize}

\textit{\textbf{Evaluation Steps:}}

\begin{itemize}
    \item \textit{Read the conversation between the Examiner and the Candidate carefully.}
    \item \textit{Assign a score for GRAMMAR AND VOCABULARY and DISCOURSE MANAGEMENT on a scale of 1 to 5, where 1 is the lowest and 5 is the highest.}
    \item \textit{Please disregard the response provided by ``Examiner" in your evaluation.}
    \item \textit{Present the assessment criteria and scores in JSON format and name it OUTPUT and the OUTPUT will have two key-value pairs: GRAMMAR AND VOCABULARY, and DISCOURSE MANAGEMENT}
\end{itemize}

\textit{\textbf{Conversation:} \{conversation\}} 

\vspace{0.2cm}

\textit{[/INST]}

\vspace{0.2cm}

\textit{\{output\}}

\vspace{0.2cm}

\textit{\textless s\textgreater }

\subsection{Instruction datapoint with performance descriptors}

\vspace{0.2cm}

\textit{\textless s\textgreater [INST]}

\vspace{0.2cm}

\textit{\textbf{Role:}}
\begin{itemize}
    \item \textit{You are an assessor of the CEFR B2 English speaking assessment. You are an expert in this field with several years of experience.}
    \item \textit{You will be given a conversation between an Examiner and a Candidate, and your task is to give scores for two metrics for the responses given by the ``Candidate" in the conversation.} 
\end{itemize}

\textit{\textbf{Evaluation Steps:}}

\begin{itemize}
    \item \textit{Read the conversation between the Examiner and the Candidate carefully.}
    \item \textit{Assign a score for the assessment criteria based on the `Performance Descriptors' given on a scale of 1 to 5, where 1 is the lowest and 5 is the highest.}
    \item \textit{Please disregard the response provided by ``Examiner" in your evaluation.}
    \item \textit{Present the assessment criteria and scores in JSON format and name it OUTPUT and the OUTPUT will have two key-value pairs: GRAMMAR AND VOCABULARY, and DISCOURSE MANAGEMENT}
\end{itemize}

\textit{\textbf{Performance Descriptors}}

\begin{itemize}
\item \textit{GRAMMAR AND VOCABULARY: \{performance descriptor grammar and vocabulary\}}
\item \textit{DISCOURSE MANAGEMENT: \{performance descriptor for discourse management\}}
\end{itemize}

\textit{\textbf{Conversation:} \{conversation\}} 

\vspace{0.2cm}

\textit{[/INST]}

\vspace{0.2cm}

\textit{\{output\}}

\vspace{0.2cm}

\textit{\textless s\textgreater }

\subsection{Instruction datapoint with evaluations steps and output format}

\vspace{0.2cm}

\textit{\textless s\textgreater [INST]}

\vspace{0.2cm}

\textit{\textbf{Evaluation Steps:}}

\begin{itemize}
    \item \textit{Read the conversation attentively, focusing on the interaction.}
    \item \textit{Rate the Candidate's use of GRAMMAR AND VOCABULARY on a scale from 1 to 5.}
    \item \textit{Rate the Candidate's ability in DISCOURSE MANAGEMENT on the same scale.}
    \item \textit{Only assess the Candidate's responses, ignoring the Examiner's input.}
    \item \textit{Present your evaluation scores in a JSON format named OUTPUT, with two key-value pairs: GRAMMAR AND VOCABULARY and DISCOURSE MANAGEMENT.}
\end{itemize}

\textit{\textbf{Example of OUTPUT format:} \{output\_format\}} 

\vspace{0.2cm}

\textit{\textbf{Conversation:} \{conversation\}} 

\vspace{0.2cm}

\textit{[/INST]}

\vspace{0.2cm}

\textit{\{output\}}

\vspace{0.2cm}

\textit{\textless s\textgreater }

\section{Evaluation prompts}\label{appendix_evaluation}

The two sets of prompts that were used to evaluate the performance of state-of-the-art LLMs without LoRA and `\textit{EvalYaks}' for part 1 are given below.

\subsection{Prompt without performance descriptors}

\vspace{0.2cm}

\textit{\textbf{Role:}}
\begin{itemize}
    \item \textit{You are an assessor of the CEFR B2 English speaking assessment. You are an expert in this field with several years of experience.}
    \item \textit{You will be given a conversation between an Examiner and a Candidate, and your task is to give scores for two metrics for the responses given by the ``Candidate" in the conversation.} 
\end{itemize}

\textit{\textbf{Evaluation Steps:}}

\begin{itemize}
    \item \textit{Read the conversation between the Examiner and the Candidate carefully.}
    \item \textit{Assign a score for GRAMMAR AND VOCABULARY and DISCOURSE MANAGEMENT on a scale of 1 to 5, where 1 is the lowest and 5 is the highest.}
    \item \textit{Please disregard the response provided by ``Examiner" in your evaluation.}
    \item \textit{Present the assessment criteria and scores in JSON format and name it OUTPUT and the OUTPUT will have two key-value pairs: GRAMMAR AND VOCABULARY, and DISCOURSE MANAGEMENT}
\end{itemize}

\textit{\textbf{Conversation:} \{conversation\}}

\subsection{Prompt with performance descriptors}

\vspace{0.2cm}

\textit{\textbf{Role:}}
\begin{itemize}
    \item \textit{You are an assessor of the CEFR B2 English speaking assessment. You are an expert in this field with several years of experience.}
    \item \textit{You will be given a conversation between an Examiner and a Candidate, and your task is to give scores for two metrics for the responses given by the ``Candidate" in the conversation.} 
\end{itemize}

\textit{\textbf{Evaluation Steps:}}

\begin{itemize}
    \item \textit{Read the conversation between the Examiner and the Candidate carefully.}
    \item \textit{Assign a score for the assessment criteria based on the `Performance Descriptors' given on a scale of 1 to 5, where 1 is the lowest and 5 is the highest.}
    \item \textit{Please disregard the response provided by ``Examiner" in your evaluation.}
    \item \textit{Present the assessment criteria and scores in JSON format and name it OUTPUT and the OUTPUT will have two key-value pairs: GRAMMAR AND VOCABULARY and  DISCOURSE MANAGEMENT}
\end{itemize}

\textit{\textbf{Performance Descriptors}}

\begin{itemize}
\item \textit{GRAMMAR AND VOCABULARY: \{performance descriptor grammar and vocabulary\}}
\item \textit{DISCOURSE MANAGEMENT: \{performance descriptor for discourse management\}}
\end{itemize}

\textit{\textbf{Conversation:} \{conversation\}}


\bibliographystyle{elsarticle-num}

\bibliography{elsarticle-num}


\end{document}